\documentclass[conference]{IEEEtran}
\usepackage{times}

\usepackage[sort,numbers]{natbib}
\usepackage{multicol}
\usepackage{epsfig}
\usepackage{graphicx}
\usepackage{amsmath,bm}
\usepackage{amssymb}
\usepackage{amsfonts}
\usepackage{bbold} 
\usepackage{array}
\usepackage{tabularx}
\usepackage{siunitx}
\usepackage{threeparttable}
\usepackage{booktabs,caption}
\usepackage{graphbox} 
\usepackage{paralist}
\usepackage{color}
\usepackage{multirow}
\usepackage[export]{adjustbox}

\usepackage{algorithm}
\usepackage[noend]{algpseudocode}
\usepackage{array,multirow}

\usepackage[pagebackref=true,breaklinks=true,colorlinks,bookmarks=false]{hyperref}

\newcommand\blfootnote[1]{%
  \begingroup
  \renewcommand\thefootnote{}\footnote{#1}%
  \addtocounter{footnote}{-1}%
  \endgroup
}

\definecolor{mygreen}{rgb}{0, 0.5, 0.02}

\usepackage[colorinlistoftodos]{todonotes}

\newcommand{\RV}[1]{\textcolor{black}{#1}}

\title{{\Large \bf Real-time Soft Body 3D Proprioception via Deep Vision-based Sensing}}

\begin{document}


\author{
Ruoyu Wang$^{1}$, Shiheng Wang${^{1*}}$, Songyu Du${^{1*}}$,  Erdong Xiao${^{1}}$, Wenzhen Yuan$^{2}$, Chen Feng${^{1\dagger}}$
}

\twocolumn[{%
\renewcommand\twocolumn[1][]{#1}%
\maketitle
\vspace{-8mm}
\begin{center}
    \centering
    \begin{minipage}{.52\textwidth}
        \centering
        \includegraphics[width=0.2\textwidth]{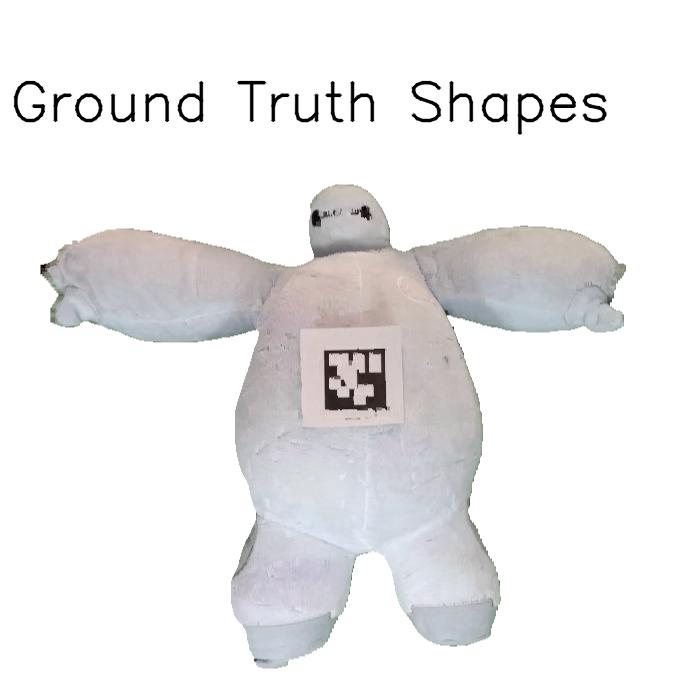}%
        \includegraphics[width=0.2\textwidth]{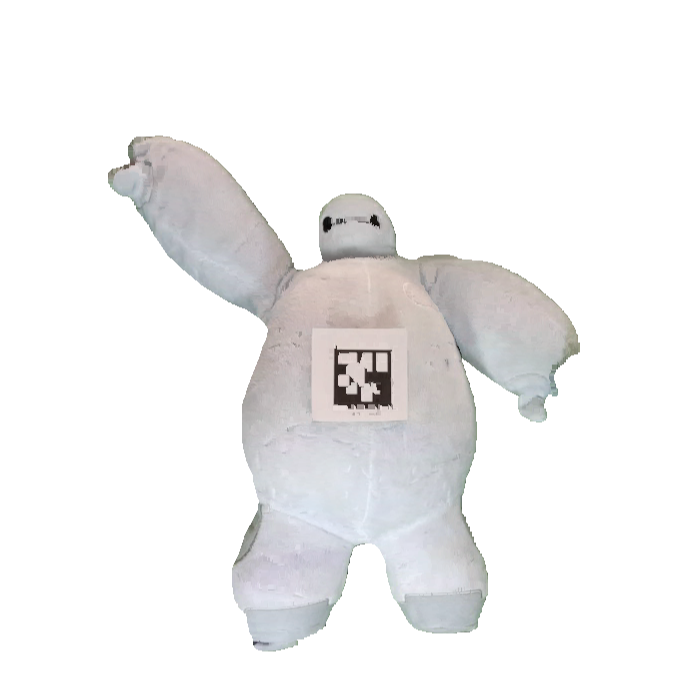}%
        \includegraphics[width=0.2\textwidth]{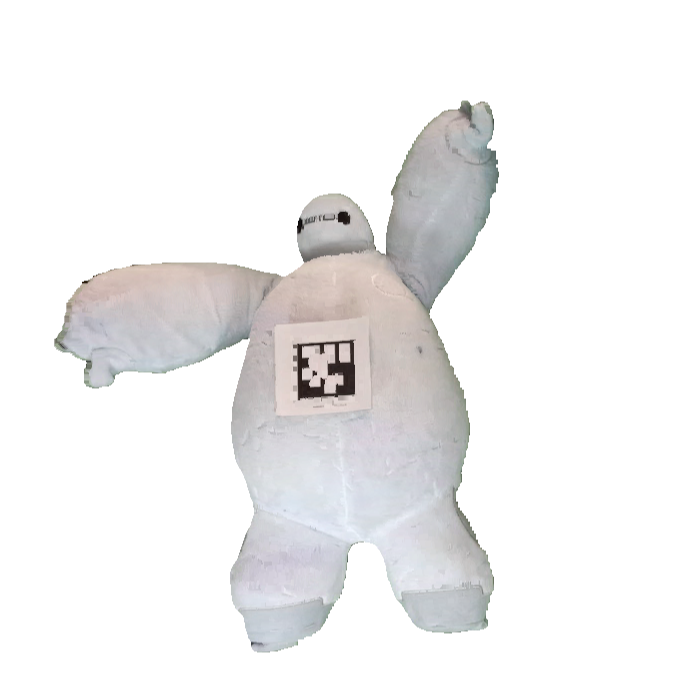}%
        \includegraphics[width=0.2\textwidth]{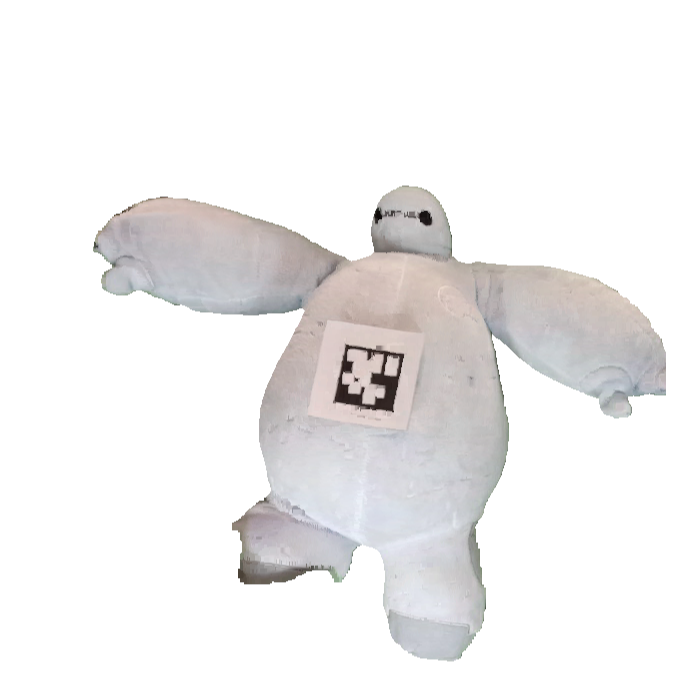}%
        \includegraphics[width=0.2\textwidth]{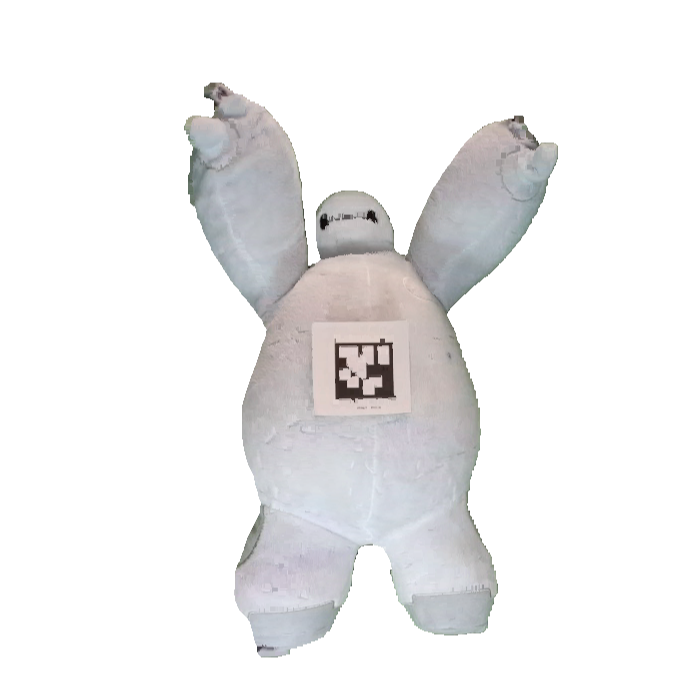}%
        \\
        \includegraphics[width=0.2\textwidth]{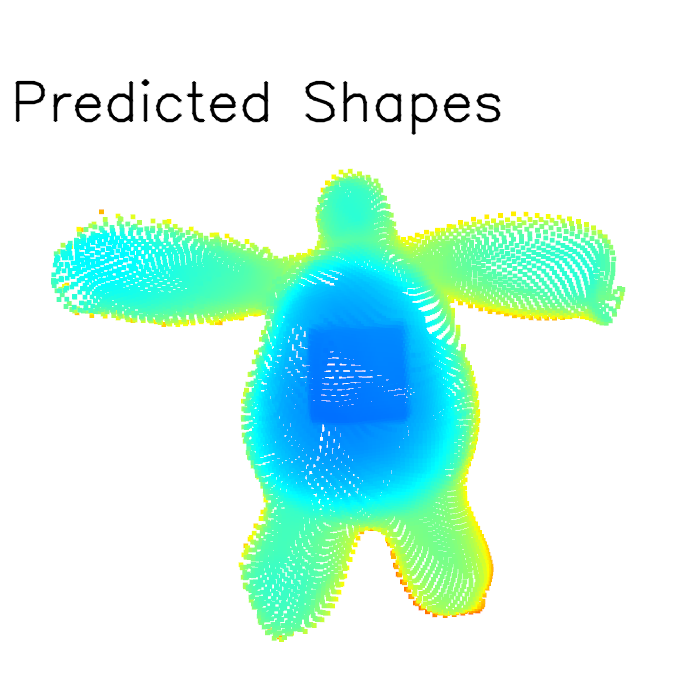}%
        \includegraphics[width=0.2\textwidth]{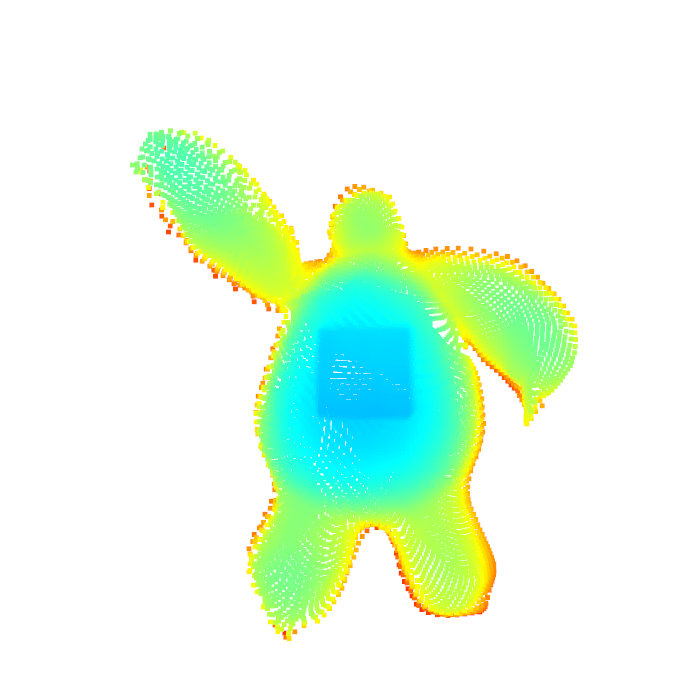}%
        \includegraphics[width=0.2\textwidth]{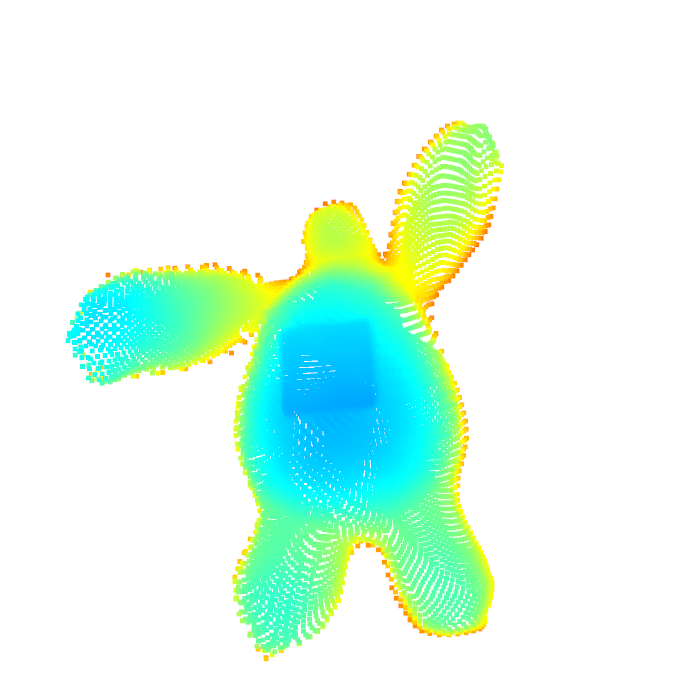}%
        \includegraphics[width=0.2\textwidth]{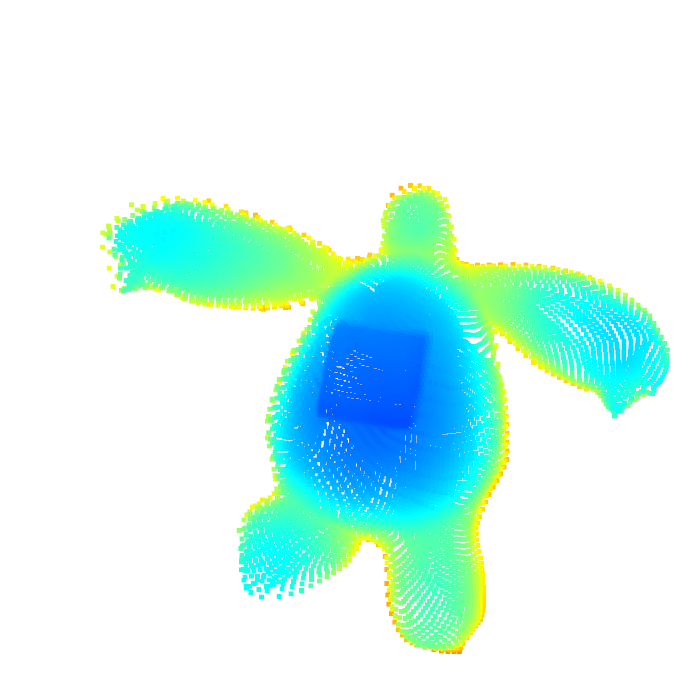}%
        \includegraphics[width=0.2\textwidth]{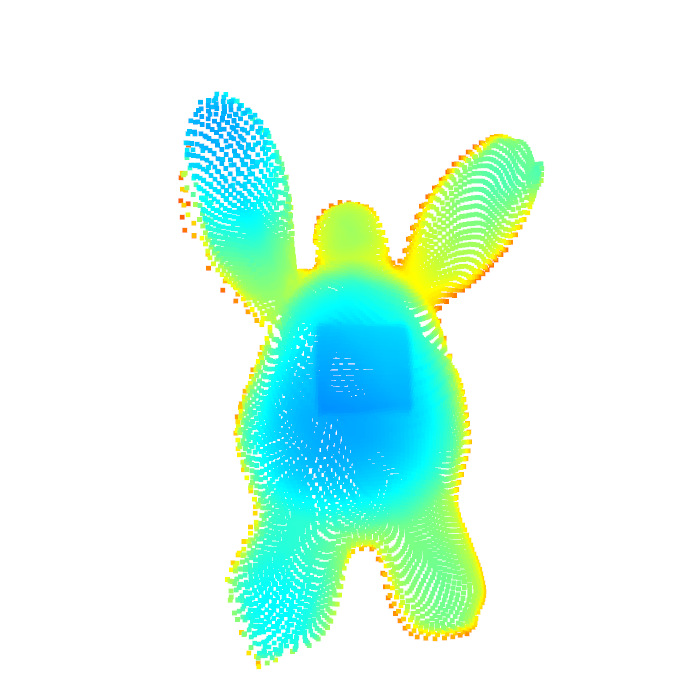}%
        
        \footnotesize{(a) 3D shape proprioception results.}
    \end{minipage}%
    \begin{minipage}{.48\textwidth}
        \centering
        \includegraphics[width=1\textwidth]{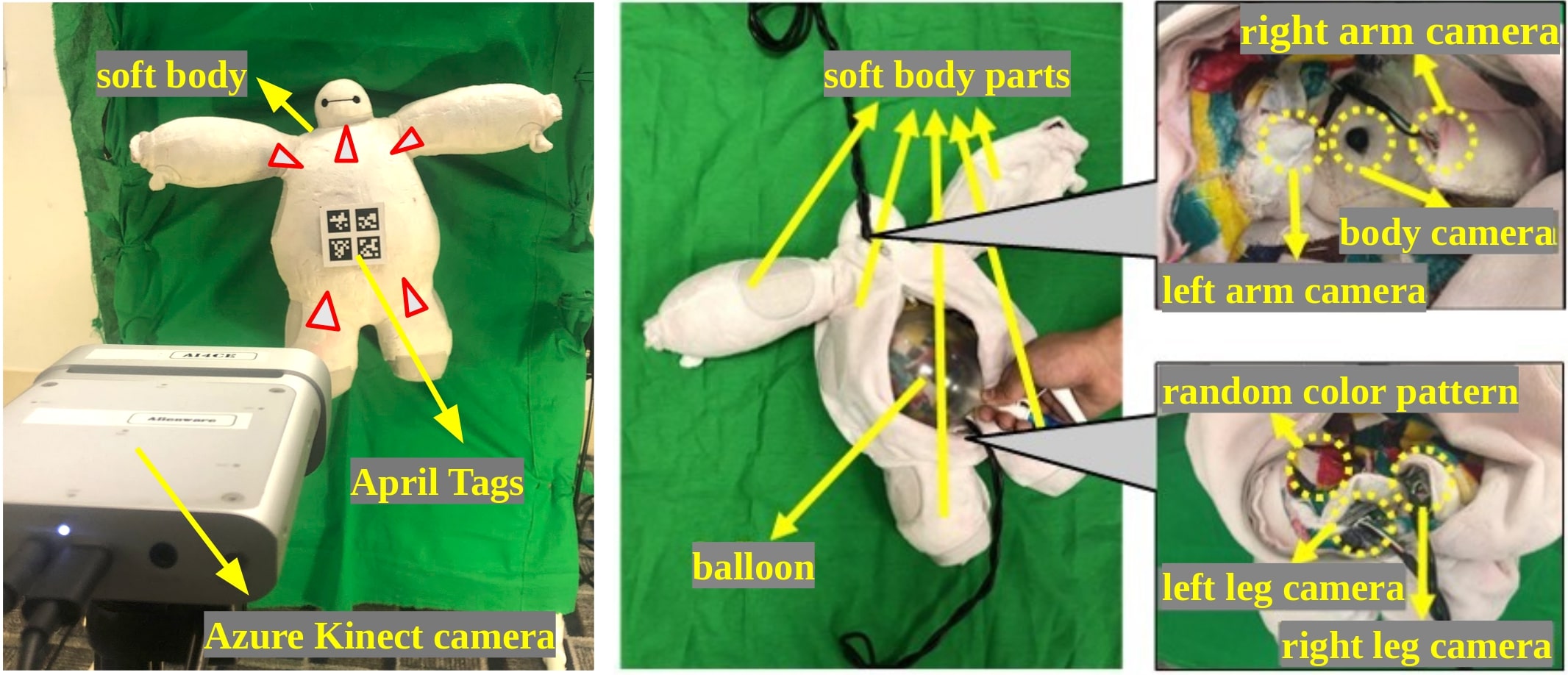}
        \footnotesize{(b) Sensing and data collection system.}
    \end{minipage}%
\end{center}%
\refstepcounter{figure}\normalfont\footnotesize{Fig. \thefigure.~~ 
Proprioception of a Baymax-shaped soft body.  The \textit{predicted 3D shapes} (point clouds in the bottom row) in (a) are inferred \textit{solely} via RGB images from cameras embedded \textit{inside} the soft body in (b) (red triangles). The point colors \RV{indicate} depth increasing from blue to red. The \textit{ground truth 3D shapes} (top row) in (a) are captured by an RGBD camera (Kinect) in (b). The predicted 3D shapes align well with the ground truth.}
\label{fig:fig1}
\vspace{3mm}
}]

\blfootnote{$^1$~New York University, Brooklyn, NY 11201.}
\blfootnote{\texttt{\{rw2380,sw4018,sd3231,edxiao,cfeng\}@nyu.edu}}
\blfootnote{$^2$~Carnegie Mellon University, Pittsburgh, PA 15213. \texttt{yuanwz@cmu.edu}}
\blfootnote{$^*$ Equal contributions.}
\blfootnote{$^{\dagger}$~The corresponding author is Chen Feng. \texttt{ cfeng@nyu.edu}}

\vspace{-3mm}
\begin{abstract}
Soft bodies made from flexible and deformable materials are popular in many robotics applications, but their proprioceptive sensing has been a long-standing challenge. In other words, there has hardly been a method to measure and model the high-dimensional 3D shapes of soft bodies with internal sensors. 
We propose a framework to measure the high-resolution 3D shapes of soft bodies in real-time with embedded cameras. The cameras capture visual patterns inside a soft body, and a convolutional neural network (CNN) produces a latent code representing the deformation state, which can then be used to reconstruct the body's 3D shape using another neural network. We test the framework on various soft bodies, such as a Baymax-shaped toy, a latex balloon, and some soft robot fingers, and achieve real-time computation ($\leq$2.5ms/frame) for robust shape estimation with high precision ($\leq$1\% relative error) and high resolution. We believe the method could be applied to soft robotics and human-robot interaction for proprioceptive shape sensing.
\end{abstract}
\IEEEpeerreviewmaketitle

\section{Introduction}
\label{sec:intro}

Soft bodies are deformable objects made from flexible and soft materials or compliant mechanisms that have very high degree-of-freedoms in their configurations. They are often used to build soft robots or human-robot interaction interfaces. 
In particular, thanks to the compliance and flexibility in soft bodies, soft robots can passively yield to the external physical contact, which makes them safe in contact-rich tasks. The infinite dimension of their shapes also enables them to fit complicated environments. 
Therefore, they show significant potentials to be applied to safety-related tasks, dexterous grasping and manipulation, and surgical applications.  
However, the high flexibility of soft bodies poses extra challenges on their proprioception, where the core tasks are to perceive their shapes in real-time. Compared to their rigid peers, soft robots are usually highly under-actuated and highly nonlinear, and external loads or contact will cause prominent deformation of the robots. 
The high-dimensional deformation can hardly be fully measured by traditional sensors, and representing the high-dimensional shapes is also challenging. 
Without proper measurement and representation of soft bodies, it is hard to perform closed-loop control on those robots.

Most of the present techniques for soft body proprioception \RV{are} based on measuring local deformation with stretchable sensors ~\cite{harnett2016flexible, jin2018wish, wessely2018shape}. 
However, the spatial sparsity design of those sensors can hardly measure the high-dimensional deformation with the desired accuracy, and the manufacturing and the transformation to other soft robots are big challenges. 
Moreover, representing the high-dimensional 3D shapes of soft robots is challenging. A traditional practice is to use Finite Element Analysis~\cite{runge2017framework, zhang2016kinematic}, but these methods have many constrains on the application conditions and \RV{require} huge computational resources, making it intractable to run in real-time.  

We hereby propose a framework to measure and represent the real-time 3D deformation of soft bodies using a data-driven approach. 
We paint random patterns inside/outside the soft bodies, embed cameras on the soft bodies to observe those patterns under various deformations. We then train a Convolutional Neural Network (CNN) to encode the visual signals to a latent space for datasets involving complex motions, and send the latent representation vectors to a decoding neural network to reconstruct the soft bodies' 3D shapes.

We test our method on multiple soft bodies/robots: a Baymax-shape toy, a latex balloon,  an elastomeric origami \cite{martinez2012elastomeric}, a PneuNets \cite{mosadegh2014pneumatic} and a fiber reinforcement actuator \cite{polygerinos2015modeling}. 
We collect datasets of more than 5,000 instances for each soft body, which contain either active free deformation or deformation caused by external force or contact. The datasets are used to train the neural networks for learning the representations of the soft bodies' 3D shapes. This training can be viewed as a ``sensor calibration'' process, where the ground truth 3D shapes are obtained by low-cost RGBD cameras. 
Our experiments (see Figure~\ref{fig:fig1}) demonstrate our method's high 3D shape sensing accuracy (absolute error: $\approx$1 \si{mm}, relative error: $\leq$1\%, when the reconstruction resolution is 100 $\times$ 100 points) in real-time computational speed ($\geq$400Hz).

To our knowledge, the proposed system is the first real-time vision-based system capable of measuring 3D shapes of soft bodies without external sensors. The accuracy and spatial resolution of the measurement significantly surpass the ones using traditional methods. 
Our method runs in real-time on GPU and could be used in the future for closed-loop control in complicated environments and motion planning for tasks that require more precise motion. We further note that the learning part of our method is barely related to the sensor design, thus can be readily transformed to other soft bodies that may not fit for camera-based sensing due to heavy occlusions.

\section{Related Work}
\label{sec:related}


\textbf{Soft robot proprioception.}
Traditionally, researchers embed stretchable sensors in the soft robots to measure local deformation. Those sensors include capacitive or resistive sensors that \RV{provide a response which is proportional to their deformation}, and optical fibers whose light conductivity decreases during bending. 
~\citet{glauser2018deformation} use a capacitive sensor embedded in a soft robot and a multi-layer perceptron (MLP) to predict 3D positions of a set of key points from the capacitive readings.  ~\citet{van2018soft} model a soft foam's state as a 2D vector of bend/twist angles, and use 30 optical fiber readings and K-Nearest-Neighbors/SVM/MLP models to learn the angles. ~\citet{molnar2018optical} also use optical fibers as input to a 2-layer MLP for estimating the end-effector's 3D position of a linear soft pneumatic robot. \RV{\citet{thuruthel2019soft} use cPDM sensors to predict the 3D position of a soft finger tip via a Recurrent Neural Network (RNN). Similarly, \cite{scimeca2019model} employ capacitive tactile sensor array and MLP to predict finger tip positions, and then reconstruct the finger's 3D shape via interpolation, although their full body shape accuracy is not evaluated.}
For a more comprehensive summary of the existing measurement methods, we refer readers to ~\cite{wang2018toward}. 
A major limitation of those methods is that they are ``over-simplified''~\cite{wang2018toward} while trying to model the high-dimensional shapes of the soft robots with low-dimensional vectors, thus \RV{compromising} the accuracy and spatial resolution. The complicated driving conditions, where the robots are driven by multiple loads, will be challenging for those sensors. This is caused by the intrinsic low resolution of the sensor components.
Our method employs vision-based designs, which offer high-resolution information regarding the high-dimensional deformation of the soft robots. The deep neural network models we developed then \RV{turn} the raw readings into a full description of the 3D shapes of the robots. Our measuring method has low dependency to the robot design or the loading conditions, hence can be widely applied on different soft robots and to complicated working environments.

\textbf{Vision-based sensors for soft bodies.}
Vision-based sensors have been designed in other areas to measure the shape of the soft bodies, like robot tactile sensing. Those sensors~\cite{Ferrier2000,GelForce2005,chorley2010tactile,yamaguchi2017implementing} use a piece of soft material as the sensing medium, with some dot patterns painted on the surface or in the body, then use an embedded camera to track the motion of the dots. Most of those sensors aim at measuring the contact force from the deformation. \citet{yuan2017gelsight} introduce a similar sensor, but they also use the reflection from the soft material surface to reconstruct the high-resolution shape of the soft body. The successful practice of those sensors shows that vision-based sensor offers a convenient way to measure the deformation of soft materials, but \RV{those sensors are not suitable for} the soft bodies that are studied in this work. The deformation of the soft material is much larger and more complex in this work \RV{than what those sensors are able to measure.}

\begin{figure*}[!t]
    \centering
    \includegraphics[width=1\textwidth]{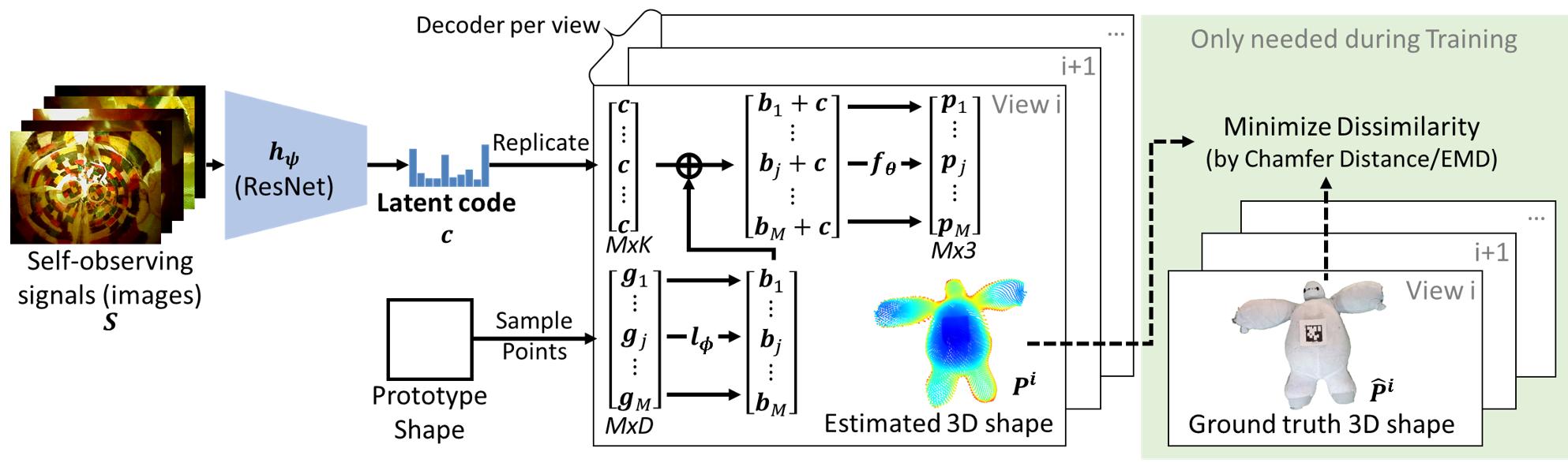}
    \caption{\textbf{Algorithm pipeline}. Our network enables a generic soft robot perception framework. The network input is self-observing signals (in this paper, we use RGB images from the embedded \RV{cameras}), and the output is the estimated 3D shapes of the soft robot \RV{generated from a prototype 3D mesh model that resembles the staic shape of the robot.} 
    Through training, a latent representation, the $K$-dimentional latent code, of the soft robot are automatically discovered, and enables a 3D shape estimation with high accuracy and arbitrary resolution.     Solid arrows in the figure indicate forward propagation, and dashed arrows means loss calculation. $\bigoplus$ means element-wise addition.
    }
    \label{fig:overview}
    \vspace{-11pt}
\end{figure*}

\textbf{3D shape reconstruction from images using neural networks.} 
The computer vision community has been working on reconstructing 3D models from single images for decades. Works like~\cite{choy20163d}, \cite{sun2018pix3d} and \cite{fan2017point} focus on reconstructing the 3D models of commonly seen rigid objects. There are also works like~\cite{alp2018densepose} to reconstruct 3D human body structure, while the shapes are partially deformable.
To reconstruct 3D shapes, up-convolutional (or transpose-convolutional) networks are often used~\cite{3dgan}. The recent FoldingNet~\cite{yang2017foldingnet} provides another more light-weight and accurate possibility to reconstruct shapes as point clouds from deep parametric surfaces, thus being particularly suitable for representing soft robots' shapes.
In this work, we also aim at reconstructing the 3D shapes from a single image using neural networks, but compared to those existing works where the images are external observation of the target object, our input data is from the internal view of the robots, and the intuitive correlation between the images and the shapes \RV{is} much weaker.

\section{Representation Learning for Soft Bodies}
\label{sec:method}


\subsection{Overview}


The proposed framework (Figure~\ref{fig:overview}) has three main steps: 

\textbf{Data collection.} Images from the embedded cameras captured at a same \RV{instant} $t$ are combined to a multi-channel image $\bm{S}_{t}$. Then a set $\mathbb{S}=\{\bm{S}_t\}$ of images at different \RV{instants} are collected, and their corresponding 3D shapes are collected \RV{with outside Kinect cameras} into a \RV{point cloud} set $\mathbb{\hat{P}}=\{\bm{\hat{P}}_t\}$. Elements in $\mathbb{S}$ and $\mathbb{\hat{P}}$ are paired up. 
 
 \textbf{Deep calibration.} A neural network is trained to model two-stage sensing functions $h_{\bm{\psi}}: \mathbb{S} \to \mathbb{H}$ and $f_{\bm{\theta}}: \mathbb{H} \to \mathbb{P}$, where $\mathbb{H}$ is the latent space inside which a vector $\bm{c}$ represents a state of the soft body. 
  $\bm{\psi}$ and $\bm{\theta}$ are learnable parameters. This process is analogous to the traditional sensor calibration with multiple inputs and outputs, so we name it as deep calibration.
  
  \textbf{Deployment.} Using the learned functions, data-driven sensing can be performed by mapping current self-observed image $\bm{S}$ to the shape $\bm{P}$ without any on-the-fly fitting as required in NURBS-based methods~\cite{zhuang2018fbg}.

With the development of deep learning, it is not difficult to find a proper encoder $h_\psi$ depending on the self-observing sensors. For instance, CNN can be used for cameras, MLP for sensor arrays, and graph convolutional networks (GCN\RV{s}) for sensor networks. However, the decoder $f_\theta$ could only be 3D up-convolutional \RV{network}~\cite{3dgan} or point cloud \RV{decoder}, because our goal is to sense the 3D shape of soft bodies, which is essential for follow-up tasks like control and motion planning for soft robots. Considering the computational efficiency for robotic applications, we propose to use point cloud decoders rather than 3D up-convolutional networks, because point cloud is a more concise representation of 3D shape, while voxels (used in 3D up-convolutional networks) naturally have the trade-off between memory/computation and shape resolution. Besides, point \RV{cloud} consists of 3D coordinates directly, which can be readily used in tasks such as motion planning for soft robots.

Among existing point cloud decoders, FoldingNet \cite{yang2017foldingnet} provides the state-of-art point cloud decoder, which supports to decode an arbitrary number of points, providing theoretically infinite 3D shape resolutions for soft bodies. Thus we choose to take advantage of FoldingNet-like 3D shape decoders.


\vspace{-3mm}

\RV{\subsection{Original FoldingNet Architecture}}
\label{sec:sec:netarch}

The original FoldingNet provides an auto-encoder architecture for point clouds. It uses PointNet\cite{qi2017pointnet} as its encoder and a share-weight MLP as its decoder. The PointNet encodes a point cloud $\bm{\hat{P}}=\{\bm{\hat{p}}_i\}$ with $N$ points into a code word $\bm{c}$ in a $K$ dimensional latent space, then the code word is replicated for $M$ times as $\mathbb{1}_M \bm{c}$ ($\mathbb{1}_M$ is a $M$-dimensional column vector with all entries being one), and concatenated to a $D$-dimension point grid $\bm{G}=[\bm{g}_j]$. $M$ can be chosen according to any desired decoding resolution even after training. After that, the $M\times{}(D+K)$ intermediate variable $[\bm{G}, \mathbb{1}_M \bm{c}]$ is fed to a share-weight MLP (weights shared across points) to obtain the decoded $M\times{}3$ point cloud as $\bm{P}=[\bm{p}_j]$. 

Mathematically, the original FoldingNet decoder realizes the following point-wise decoding function:
\begin{equation}
\bm{p}=f_{\theta}(\bm{g},\bm{c}): \mathbb{R}^D\times \mathbb{R}^K \to \mathbb{R}^3,
\label{eq:folding}
\end{equation}
where $\theta$ are learnable parameters of the share-weight MLP.
When $D$ is 2 or 1, FoldingNet can be considered as a deep parametric surface/curve as analogous to a NURBS surface/curve, where $\theta$ are knot vectors defining basis functions, $\bm{c}$ control points defining the shape of control mesh, and $\bm{g}$ the spline parameter. Varying the value of $\bm{g}$ will trace out the 3D surface/curve shape. The advantage of using FoldingNet than NURBS is that FoldingNet enables better and more flexible data-driven learning of the shape, while knot vectors in NURBS can not be optimized easily in surface fitting.

\vspace{-1mm}
\RV{\subsection{Improved FoldingNet Architecture}}

Although FoldingNet already has some desirable properties, there are still several facts in the original FoldingNet decoder that \RV{call} for improvement for more effective applications in our framework:
\begin{itemize}
  \item Because $D<<K$, $[\bm{g}_i, \bm{c}]$ is dominated by the same $K$-dimensional vector $\bm{c}$, which slows down the learning.
  \item The parameter grid $\bm{G}$ is sampled on a 2D square. However, a soft robot could potentially be of any shape. Therefore, always using a 2D square as the parameter grid could also slow down the learning, and reduce the expressive power of the network. 
  \item There is only one decoder in the original FoldingNet.
  However, it is sometimes desirable to be able to sense the complete shape of a soft body. Due to possible self-occlusions of a soft body, we need to use multiple 3D cameras to obtain the complete shape from different views.
  But multiple 3D cameras such as Kinect may interfere with each other when they operate simultaneously. One way to circumvent this interference problem is to collect data respectively from each camera with others turned off. Therefore, each individual ground truth 3D shape is only a partial view of the full shape. To get a complete 3D shape, we have to use multiple decoders.
\end{itemize}

\begin{figure}[!t]
    \centering
    \includegraphics[width=0.9\columnwidth]{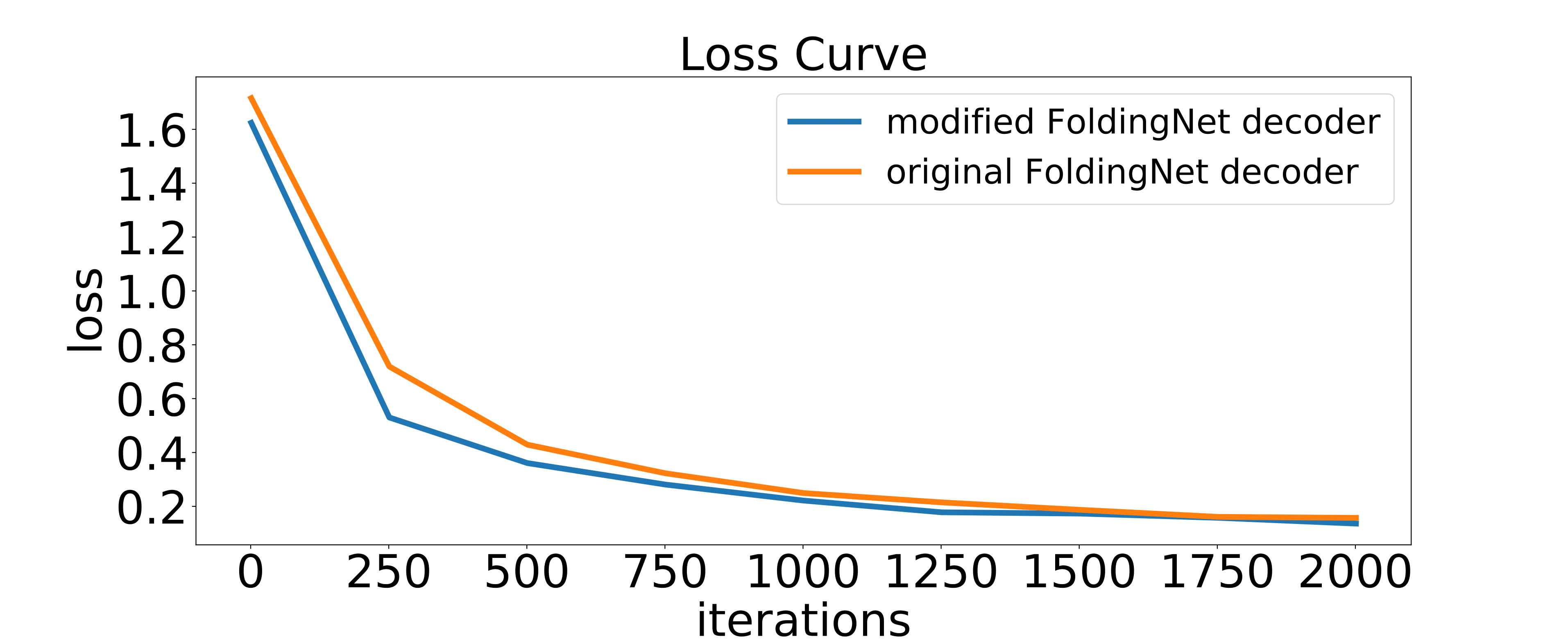}
    \caption{Loss curves of the original FoldingNet decoder and our modified one on the same training set shows that the modified decoder learns faster. Note that the original decoder has 1.1 million parameters, while the modified only has 0.5 million.}
    \label{fig:loss}
\end{figure}

Due to the above reasons, we propose to improved FoldingNet decoder formally from equation~\ref{eq:folding} to the following:
\begin{equation}
    \bm{p} = f_{\theta}(\bm{g},\bm{c}) = f_{\theta}(l_{\phi}(\bm{g})+\bm{c}),
\end{equation}
realizing the following improvements (see Figure~\ref{fig:overview}):

\textbf{Learned constant biasing.}  Instead of concatenating the grid points $\bm{G}$ to $\mathbb{1}_M \bm{c}$, each $D$-dimension grid point in $\bm{g}_j$ is first mapped to $K$-dimension as a learned constant bias $\bm{b}_j:=l_{\phi}(\bm{g}_j) \in \mathbb{R}^K$ using another share-weight MLP $l_{\phi}~:~\mathbb{R}^D~\to~\mathbb{R}^K$, then added with $\bm{c}$. This will help the training process converge faster and enable a smaller network with with less parameters. As shown in Figure~\ref{fig:loss}, this modification leads to a lighter weight network with 50\% parameter reduction yet still converges faster.

\textbf{Deforming from prototype.} Instead of always sample from a 2D square for the parameter grid, we propose to use a prototype shape, and increase the dimension of $D$ from 2 to 3. The prototype shape here is a 3D mesh model that is close to the static shape of the soft robot at its zero-state. It could be a cylinder, cube, or a rough CAD model of the robot.
    
\textbf{Multiple decoders.} Instead of using only one decoder, we enable the use of multiple different decoders (each with its own learnable parameters $\theta, \phi$). \RV{The number of decoders depends on how many portions of the soft body are expected to be predicted.} Each decoder corresponds to a particular view and is responsible to predict the partial 3D shape observed by the 3D camera from that view, given the same latent code $\bm{c}$.
In the training process, if the point cloud comes from a particular depth camera, then only the encoder weights and the weights of the corresponding decoder are updated.
Due to this asynchronous multi-view ground truth, the original Chamfer Distance ($D_{\text{Chamfer}}\{\cdot,\cdot\}$) based loss~\cite{fan2017point} has to be modified as follows.
If there are $C$ views, and $C_i$ point clouds collected from the $i$-th view, our loss function is:
\begin{equation}
\begin{aligned}
L(\psi,&\{\theta_i,\phi_i\})=\\
& \sum_{i=1}^{C}\frac{1}{C_i}\sum_{j=1}^{C_i}D_{\text{Chamfer}}\{\hat{\bm{P}_j^i}, \bm{P}_j^i(\psi,\theta_i,\phi_i)\}, 
\end{aligned}
\end{equation}
where $\bm{P}_j^i$ is the $j$-th predicted point cloud of the $i$-th view, and $\hat{\bm{P}_j^i}$ is the corresponding ground truth point cloud. $\psi$ are the parameters of encoder $h_{\psi}$, and $\theta_i,\phi_i$ are the parameters of the decoder MLPs $f_{\theta}$ and $l_{\phi}$ corresponding to the $i$-th view. The Chamfer distance~\cite{fan2017point} between point cloud $\bm{P}_a$ and $\bm{P}_b$ is: 
\begin{equation}
\begin{aligned}
D_{\text{Chamfer}}(&\bm{P}_a , \bm{P}_b) =  \frac{1}{2N_a}\sum_{\mathbf{a}\in\bm{P}_a}\min_{\mathbf{b}\in\bm{P}_b}\|\mathbf{a}-\mathbf{b}\| \\&+\frac{1}{2N_b}\sum_{\mathbf{b}\in\bm{P}_b}\min_{\mathbf{a}\in\bm{P}_a}\|\mathbf{b}-\mathbf{a}\|,
\end{aligned}
\end{equation}
where $N_a$ and $N_b$ are numbers of points in point clouds $\bm{P}_a$ and $\bm{P}_b$ respectively. 

\section{System design for our method}

Our data collection system is shown in Figure \ref{fig:fig1} (b). In this section, we discuss how we collect ground truth 3D shapes and self-observing images.
Note that our system supports using single or multiple Kinect cameras to collect ground truth. Self-observing cameras and Kinect cameras are synchronized via LCM\cite{huang2010lcm}, so that the self-observing images and ground truth shapes can be paired up.
We also discuss the fabrication of self-observing soft bodies and 
highlight the issues we addressed for data collection.

\begin{figure}[!t]
    \centering
    \includegraphics[width=0.9\columnwidth]{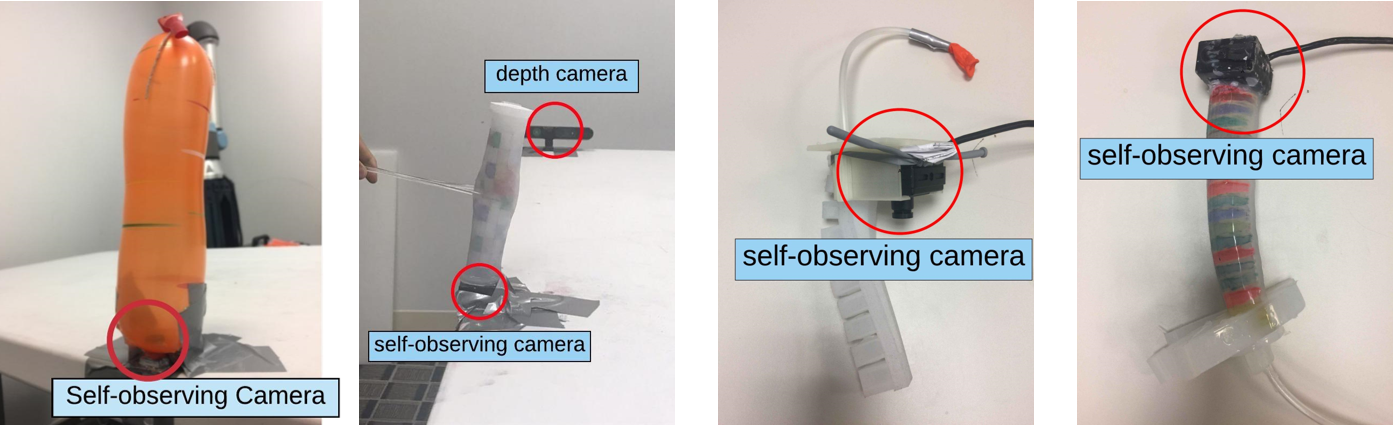} \\
    \vspace{-4pt}
    \caption{\textbf{More soft bodies.} Left to right: latex balloon, origami soft robot\cite{martinez2012elastomeric}, PneuNets\cite{mosadegh2014pneumatic}, fiber-reinforced actuator\cite{polygerinos2015modeling}.}
    \label{fig:other_robots} 
    \vspace{-5mm}
\end{figure}

\label{sec:system}
\subsection{Collecting Self-observing Images}

Figure~\ref{fig:fig1} (b) (right) shows how we collect self-observing images using cameras inside the Baymax-shaped toy. This soft body has 5 \RV{parts}: the body and the left/right arms/legs. For each \RV{part}, a self-observing camera is installed. The inner surface of the \RV{parts} are randomly colored to provide better visual cues, and the \RV{parts} are filled with inflated transparent balloons. The cameras are fixed inside small bags to restrict their movement inside the soft body.

Figure~\ref{fig:other_robots} shows the installation of self observing cameras for other soft bodies (including the spherical balloon in Figure~\ref{fig:qualitative_acc}), which contains only one self-observing camera respectively. These cameras are fixed to avoid moving along with the soft body during data collection.

\subsection{Collecting Ground Truth 3D Shapes}
Figure~\ref{fig:fig1} (b) (left) depicts how we collect ground truth 3D shapes. One or more Kinect cameras are fixed around the soft body. The soft body is fixed against a green curtain for easier background removal, as inspired by~\citet{schulman2013tracking}.

\textbf{Multiple Kinect interference.} Some models of Kinect camera will interfere with each other when they work simultaneously, as mentioned in section~\ref{sec:sec:netarch}. In this case, we collect the ground truth point clouds for each Kinect camera view separately, and apply a multi-decoder architecture.

\textbf{Multiple Kinect registration.} The relative poses of Kinect cameras have to be obtained in order to register the partial views into one common frame. We achieved this by performing Iterative Closest Point (ICP) to the point clouds collected by different Kinect cameras on a planar target.


\textbf{Common body frame.}
As~\citet{glauser2019interactive} did for hand pose estimation,
our method is only responsible for estimating the deformation relative to the body frame. Thus the 6-DOF global rotation and translation parameters of the soft body should be removed by registering ground truth shapes in the dataset into a common coordinate frame.
We define this common body frame by \RV{a set of fixed AprilTags~\cite{olson2011apriltag}  (four in our quantitative experiments) on the belly of Baymax.} The point cloud for each frame was transformed to the same reference frame according the pose tracked by the AprilTag. Note that for soft bodies that have only one self-observing camera which is static to the Kinect camera, this step is not necessary since the self-observing camera defines the body frame. Any deformation relative to the self-observing camera could then be estimated via our method.

\section{Experimental Results}
\label{sec:exper}


Now we discuss the accuracy, the memory efficency and the computational speed of our method. 
Quantitative analysis was done on both the Baymax-shaped toy dataset and the spherical balloon dataset, in order to show the performance of our method on free deformation (deformation without \RV{pressure from another object}) and compliant deformation (deformation under \RV{pressure from another object}). \RV{Free deformation was performed by randomly moving the body parts of the Baymax-shaped toy, and compliant deformation was performed by randomly squeezing the spherical balloon.} Both soft bodies were manipulated by hands with color gloves for easy background subtraction. We also analyzed the influence of hyper-parameters, including the input image resolution and the latent space dimension, with respect to the 3D reconstruction error. 

\RV{Most} experiments were run 
on NVIDIA TITAN Xp GPU. \RV{Computational speed was also tested on Jetson AGX Xavier}. We use PyTorch~\cite{PytorchWebsite} to implement the neural networks.

\subsection{Experimental Settings}

\textbf{Data preprocessing.}
For a more effective representation of the neural networks, we normalized the coordinates of point clouds in the depth images, either the ground truth ones from RGBD camera, or the output of the neural networks, into $[-1, 1]$. The images from the embedded camera were down-sampled to $224\times224$.

\RV{\textbf{Network details.}
We applied ResNet18\cite{he2016deep} to implement $h_{\bm{\psi}}$ due to its high efficiency. The dimension of latent space $K$ was set to 512. $f_\theta$ was a 4-layer MLP with 512, 512, 512 and 3 neurons in each layer, and $l_\phi$ was a 3-layer MLP with 3, 256, and 512 neurons in each layer.} 

\RV{\textbf{Baseline method.}} We compared our method with a K-Nearest Neighbor  (KNN) based method. In this method, the images were down-sampled to $14\times14$ and then stretched to  vectors and mapped to a latent space using principal component analysis (PCA). The shape estimation task was considered as searching the  nearest neighbor for the input image in the latent  space, and using the corresponding  point cloud  of  the nearest neighbor as the predicted 3D shape for
the input image. The latent space dimension was the same as our method.


\textbf{Training settings.}
Training-testing division was set to 5:1. For the Baymax-shape toy dataset, we collected 5,775 samples for training and 1,225 samples for testing. For the spherical latex balloon dataset, we collected 2,666 samples for training and 534 for testing. \RV{The data were collected in four different sessions with different people manipulating the deformation of the soft bodies to reduce biases. Training and testing data were randomly selected from all collected data.}

Adam optimizer with 0.0001 learning rate was used, and the batch size was set to 16. All training processes in this paper were terminated at the 500th epoch. 

\textbf{Prototype for decoding.}
As mentioned above, using a prototype close to the shape of the soft robot as the point grid for decoding can accelerate the convergence of the training. In the experiments, a $100 \times 100$ square grid in $x$-$y$ plane, with 0 for $z$ coordinates was used as the prototype, since the point clouds of the soft bodies mainly distributed along $x$-$y$ plane.
 The points on the prototype distributed equally within $[-1, 1]$ in both $x$ and $y$ directions. 



\begin{table}
\centering
\caption{Shape Reconstruction Error (unit: \si{mm})}
\RV{
\begin{tabular}{l c c c c c c  }
\toprule
\multirow{2}{*}{$d_H$}  & \multicolumn{3}{c}{Free Deformation} & \multicolumn{3}{c}{Compliant Deformation} \\ \cline{2-7}
& Mean & Median & Max & Mean & Median & Max \\ \midrule
\textbf{Ours} & \textbf{1.2} & \textbf{0.8} & \textbf{26.0} & \textbf{7.5} & \textbf{2.6} & 96.2 \\ \midrule
\textbf{KNN} & 5.3 & 3.3 & 61.2 & 15.4 & 10.3 & \textbf{85.4}\\
\bottomrule
\end{tabular}
}

\label{Tab:accuracy}
\vspace{-5mm}
\end{table}

\subsection{Shape Reconstruction \RV{Error}}
\label{subsec:acc}
\begin{figure*}[!h]
    \centering
    \includegraphics[width=0.88\textwidth]
    {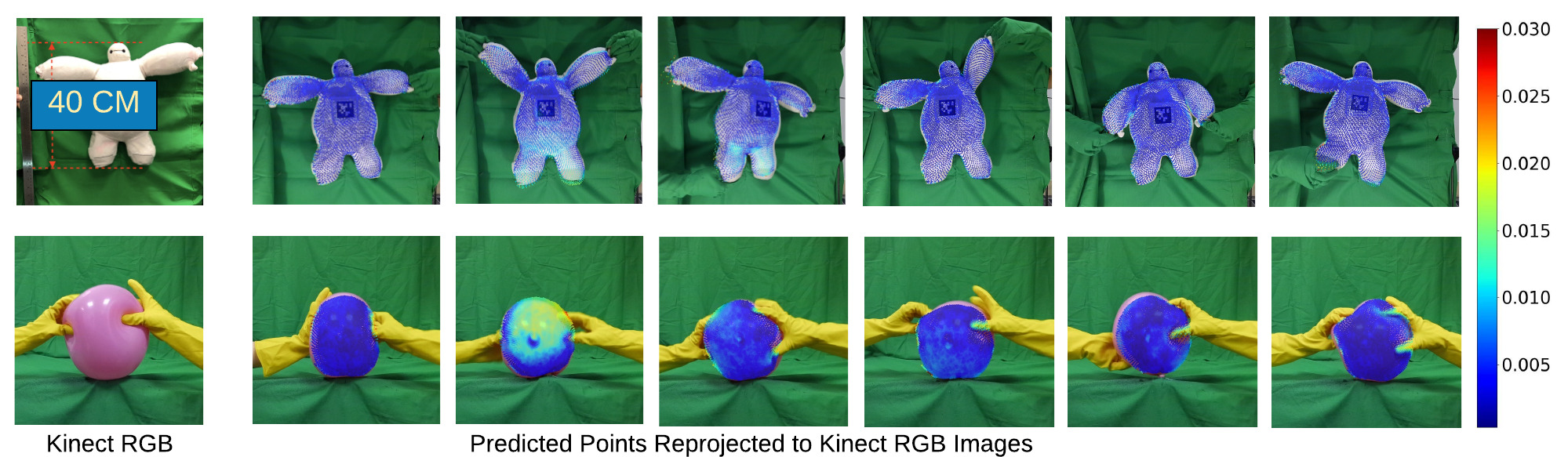}
        \caption{\textbf{Qualitative results for 3D reconstruction accuracy}. The most left column shows the original Kinect RGB images, and the other column shows the reprojection of predicted points onto the Kinect RGB images. Each point is color-coded with its distance to \RV{the closest point in ground truth point cloud (unit: \si{mm})} The top row shows the qualitative results of \textit{free deformation} (Baymax) and  the bottom row shows the qualitative results of \textit{compliant deformation} (spherical balloon).}
    \label{fig:qualitative_acc}
    \vspace{-3mm}
\end{figure*}

\RV{The shape reconstruction error was evaluated using Hausdorff distance between prediction and ground truth: 
\begin{equation}
    d_H(\hat{P}, P) =  \max\{\max_{\hat{p}\in\hat{P}}{\min_{\vphantom{\hat{p}}p\in \vphantom{\hat{P}}P}\|\hat{p}-p\|},\,\max_{\vphantom{\hat{p}}p\in \vphantom{\hat{P}}P}{\min_{\hat{p}\in \hat{P}}\|\hat{p}-p\|}\},
\end{equation} 
where $\hat{P}$ and $P$ are predicted and ground truth point clouds respectively. $\hat{p}$ and $p$ are points from $\hat{P}$ and $P$ respectively. Our evaluation metric ensures that the error is \textit{not averaged} among points in a point cloud, and it captures \textit{the worst} case between a predicted point cloud and a ground truth point cloud.}
Table~\ref{Tab:accuracy} shows errors \RV{of} shape reconstruction on the testing datasets Our method reduces the mean error  by \RV{77.3}\% and \RV{51.3}\% compared to KNN-based method on each dataset respectively. \RV{The relative mean error reaches 0.3\% and 3\%, divided by the height of the Baymax toy (400 \si{mm}) and the diameter of the balloon (250 \si{mm}).}  The results indicate that our method performs well on estimating both free and compliant deformation. Figure \ref{fig:qualitative_acc} depicts the qualitative results of predicted point clouds vs. the ground truth point clouds.

\RV{We note that our method performs better on free deformation than compliant deformation. Yet, our max error on compliant deformation is higher than that of KNN-based method. This is because during the data collection of compliant deformation, some points on the balloon were occluded by hands. Thus we excluded those points in the ground truth. However, the neural network still tried to interpolate these occluded points. In the evaluation, these points without direct ground truth are the major sources of large errors (Figure~\ref{fig:qualitative_acc}). KNN-based method does not interpolate the occluded points, which may alleviate this issue in our testing dataset.}

\subsection{Memory Efficiency}

 Our method stored 12 million single-precision floating-point numbers as the learned weights while the KNN-based method stored 166 million, leading to a reduction of 93\%. Moreover, KNN memorized all the training data, so the space will further grow on larger training sets.

\subsection{Computational Speed}

Since we can arbitrarily change the number of reconstructed points even after training, it is necessary to study the relationship between this number and the computational speed.
Table~\ref{Tab:speed} shows the computational speed vs. the number of points predicted. Our method reveals high efficiency. \RV{On the TITAN XP,} with 400\% increase in number of points, the speed decreases only by 2.0\%. This result shows that our method can reconstruct the 3D shape with even higher resolution, i.e., the number of points, than the ground truth without significantly sacrificing speed. \RV{And our method can run in real-time on an embedded system with the Jetson AGX Xavier.}


\begin{table}
\centering
\caption{Computational Speed (unit: \si{Hz})}
\RV{
\resizebox{0.48\textwidth}{!}{
\begin{tabular}{ l| c c c c c c}
\toprule
\textbf{Resolution*} & 10,000 & 14,400 &  1,9600 & 25,600 & 32,400 & 40,000 \\ \midrule
\textbf{TITAN Xp} & 430 & 428 &  423 & 428 & 423 & 421\\
\midrule
\textbf{Jetson AGX Xavier} & 29 & 24 &  20 & 17 & 15 & 13\\
\bottomrule
\end{tabular}
}
\begin{tablenotes}
      \small
      \item *The resolution is defined by the number of predicted points. 
    \end{tablenotes}
}
\label{Tab:speed}
\vspace{-1mm}
\end{table}




\subsection{Hyper-parameter Analysis}
We studied the choice of hyper-parameters to obtain a better understanding of our method, and to test the adaptability of our method to resource-limited situations. \RV{We found that the input image resolution and latent space dimension do not have significant influence on computational speed, so we only report their influences on 3D reconstruction error.}


\textbf{Resolution of input images}.
The original CNN's input image resolution was $224\times224$. In this experiment, images were first down-sampled to $16\times16$, $32\times32$, $64\times64$, and $128 \times 128$, before fed into the CNN. The dimension of latent space $K$ was kept at 512.  Table \ref{Tab:ImageRes} shows the quantitative and qualitative evaluation on the influence of reducing the image resolution. The model can provide \RV{accurate} prediction even when the image size reduces to only $16\times16$. This result indicates that our method can tolerate low-resolution images, which sometimes is the only choice of self-observing signals on robot, due to cost/space constraints and installation difficulty.
\begin{table}[t!]
\centering
\caption{Input Image Resolution Analysis  (unit: \si{mm})}
\RV{
\resizebox{0.48\textwidth}{!}{
\begin{tabular}{l| c c c c c}
\toprule
\textbf{Image Res.} & $16\times16$ & $32\times32$  &  $64\times64$ & $128\times128$ & $224\times224$ \\ \midrule
\textbf{Mean $d_H$} & 1.6 & 1.2  & 1.1  & 1.2 & 1.2 \\
\textbf{Median $d_H$} & 1.0 & 0.8 & 0.8 & 0.8 & 0.8 \\
\textbf{Max  $d_H$} & 25.7 & 39.3  &  26.6 & 27.2 & 26.0 \\ \midrule
\textbf{Example} & \includegraphics[align=c,width=0.07\textwidth]{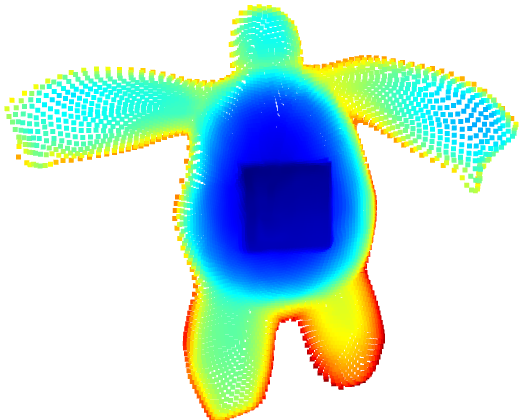} 
& \includegraphics[align=c,width=0.07\textwidth]{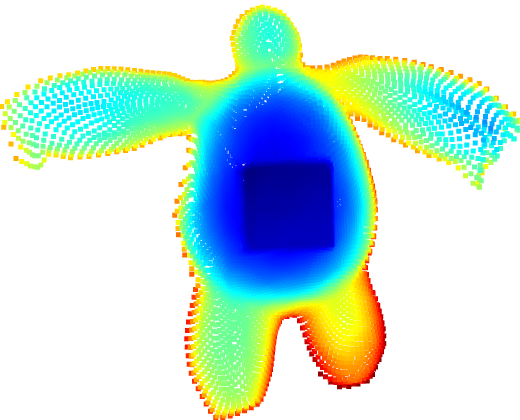} 
& \includegraphics[align=c,width=0.07\textwidth]{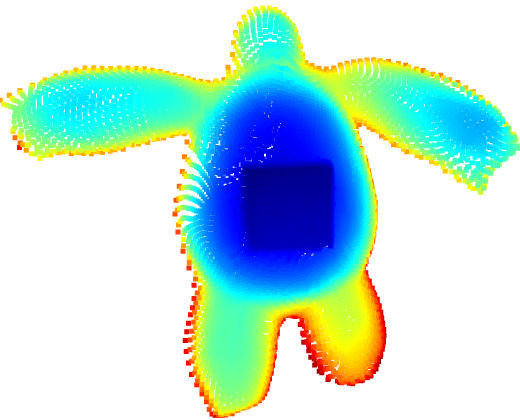} 
& \includegraphics[align=c,width=0.07\textwidth]{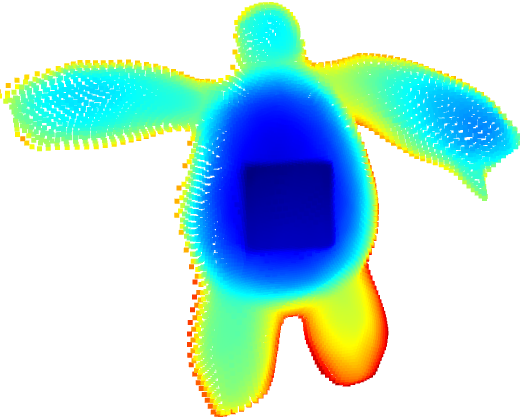} 
& \includegraphics[align=c,width=0.07\textwidth]{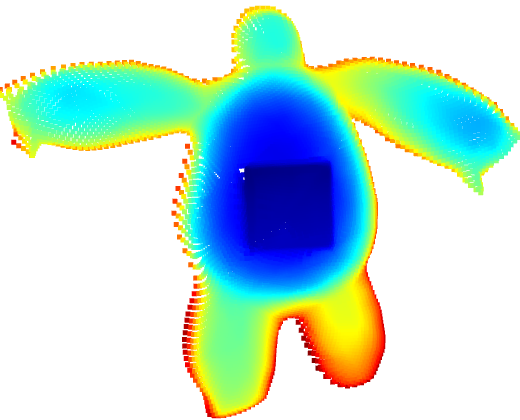}\\ \bottomrule
\end{tabular}
}
}
\label{Tab:ImageRes}
\vspace{-4mm}
\end{table}

\textbf{Dimension of latent space}.
ResNet18 can encode the input image into a 512 dimensional vector, so we first used the 512-d vector as of the latent vector. To explore if the dimension of the latent space can be further reduced, an additional fully-connected layer was added after ResNet18 to change the latent vector dimension from 512 to $K$. We experimented with the $K$ of 32, 64, 128, and 256 respectively while keeping the image size as $224\times224$. Table \ref{Tab:LatentDim}  shows the evaluation of different dimensions of the latent space. The result shows that when $K$ is lower than 64, there is a significant decrease in the prediction accuracy. This high dimensional state space is justified due to the highly nonlinear nature of soft robots, although still much more manageable than the theoretical infinite degrees of freedom of soft bodies.

\begin{table}[t!]
\centering
\caption{Latent Space Dimension Analysis (unit: \si{mm})}
\RV{
\resizebox{0.48\textwidth}{!}{
\begin{tabular}{l | c c c c c}
\toprule
\textbf{$K$} & 32 & 64  &  128  & 256 & 512 \\ \midrule
\textbf{Mean $d_H$} & 3.9 & 1.9  &  1.7 & 1.4 & 1.2 \\
\textbf{Median $d_H$} & 3.0 & 1.5  &  1.2 & 1.0 & 0.8 \\
\textbf{Max $d_H$} & 31.3 & 25.9  &  27.1 & 25.1 & 26.0 \\ \midrule
\textbf{Example} & \includegraphics[align=c,width=0.07\textwidth]{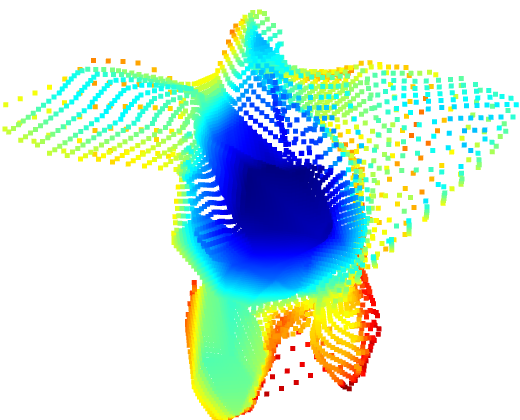} 
& \includegraphics[align=c,width=0.07\textwidth]{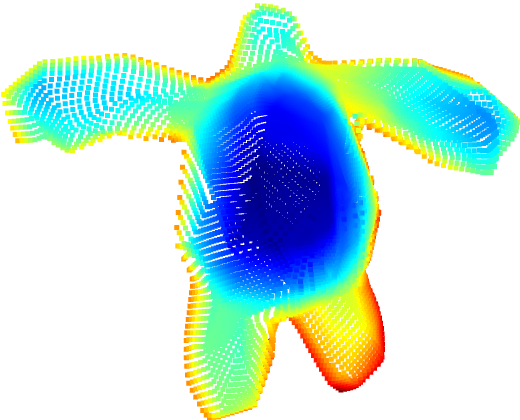} 
& \includegraphics[align=c,width=0.07\textwidth]{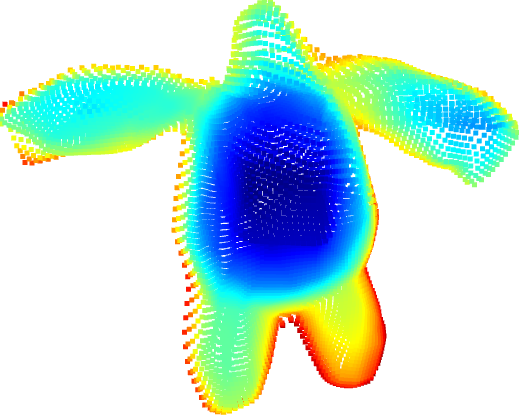} 
& \includegraphics[align=c,width=0.07\textwidth]{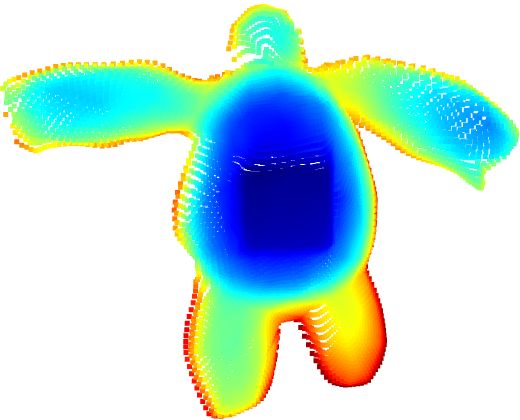} 
& \includegraphics[align=c,width=0.07\textwidth]{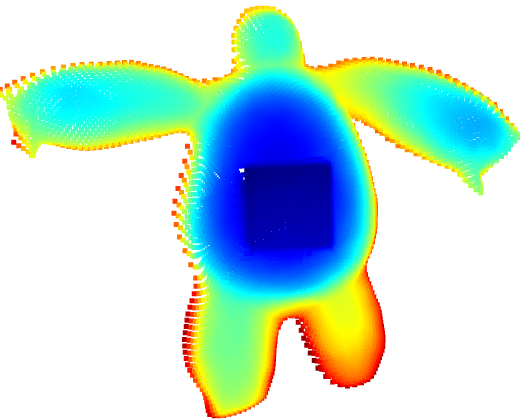}\\ \bottomrule
\end{tabular}
}
}
\label{Tab:LatentDim}
\vspace{-4mm}
\end{table}


\subsection{More Results on Other Soft Bodies}
\begin{figure}[!t]
    \centering
    \includegraphics[width=1\columnwidth]{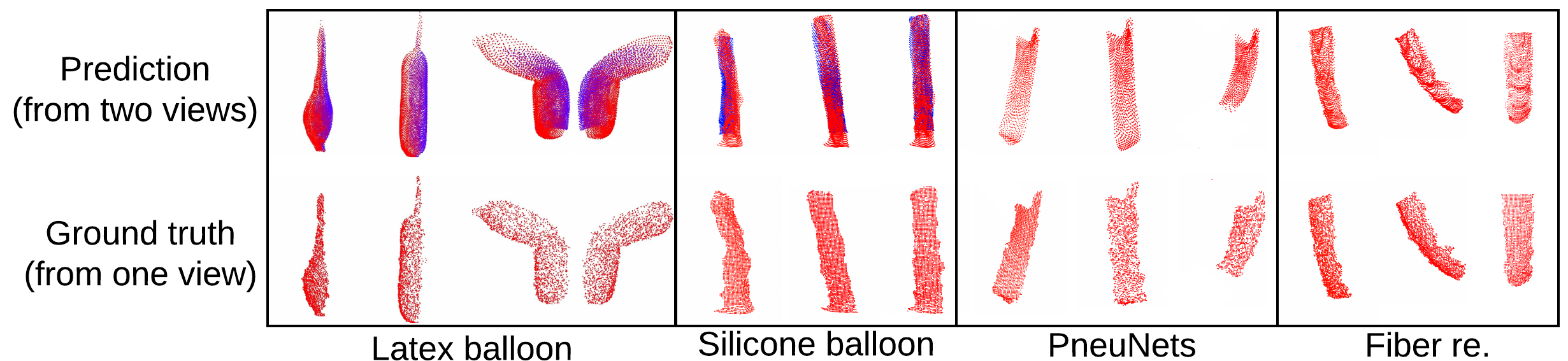}
    \caption{\textbf{More qualitative results:} From left to right: cylindrical latex balloon shrink/expand/left-bend/right-bend, origami soft robot: compliant deformation. PneuNets bend, and fiber-reinforced actuator bend. The top row illustrates our predictions and the bottom row illustrates the ground truth. Red and blue
points are predicted from two different views respectively. Although the ground truth is collected from two views separately, our multi-view decoders can predict points from two views simultaneously. \RV{Note that in these experiments the ground truth was collected from Kinect v1 with lower quality, therefore the prediction quality is not comparable to the Baymax experiment.} }
    \label{fig:more_results}
\vspace{-3mm}
\end{figure}

Here we show more qualitative results on other soft bodies, including cylindrical balloon (bending and expanding), Origami soft robot (compliant deformation), PneuNets (bending) and fiber-reinforced actuator (bending), as shown in Figure~\ref{fig:more_results}. Note that the data for \RV{cylindrical balloon and Origami soft robot were collected using two Kinect cameras separately so that multi-decoder mode is applied to these datasets}. These results show that our method can adapt to various soft bodies and deformation types.
\vspace{-2mm}

\RV{\subsection{Limitations and Future work} 
We propose a framework to measure 3D shapes of soft bodies using self-observing cameras, which works well on multiple soft bodies in the lab conditions, when the target bodies have hollow structures. In the future, we plan to improve the method's generality by adding internal lights to the soft bodies and paint their surface opaque, so that the system is robust to external lights. For non-hollow soft bodies, we will use clear materials to manufacture the soft bodies, so that the vision-based method will also be effective. For some soft robots with complicated internal structures, which we can hardly use a camera to capture the internal space of a reasonable range, vision-based methods could fail. In this case, we plan to explore using non-line-of-sight sensors to capture raw signals, and using a similar neural network method to predict the 3D shapes of the soft bodies. We will adjust the current encoder (the ResNet) according to the sensor design. 
}

\section{Conclusion and Discussion}
\label{sec:conclusion}

In this paper, we propose a vision-based sensing system to measure the soft bodies' real-time 3D shapes, which is their core proprioception and can be used for closed-loop control in precise tasks. The system uses a CNN to encode the input images from the internal cameras into latent codes, and then train a decoder neural network to reconstruct the 3D shapes of the robot from the latent codes. For the training and validation purpose, we also build a multi-Kinect system to get the ground truth shapes of the robots. Experimental results show that our system can provide an accurate and efficient measurement of both free deformation and compliant deformation of soft bodies. Compared to the existing measuring methods for soft robot \RV{proprioception}, our method well measures the dense 3D models of the robots under complicated loading conditions, and can be easily applied to different designs of the robots, especially with a minimum request in hardware fabrication. 


The neural networks we proposed for dimension reduction for the soft robot, i.e. the representation into a latent space, has more potential applications related to soft robots. For example, instead of using cameras, other embedded sensors can provide the input for the network as well. The latent code, which is a compact description of the robots' internal state, can also be used for other goals, including serving as the state for control or motion planning for soft robots. At the same time, we are interested in exploring the performance of our method on other soft robot designs. We are actively exploring these new research directions.





\addtolength{\textheight}{-0cm}   



\bibliographystyle{unsrtnat}
\bibliography{deep-learning,soft-sensing}

\end{document}